\documentclass[a4paper,fleqn]{cas-sc}

\usepackage[numbers]{natbib}
\usepackage{graphicx} 
\usepackage{lipsum} 
\usepackage{enumitem}
\usepackage[figurename=Fig.]{caption}
\usepackage{eurosym}
\setlist[itemize]{align=parleft,left=0pt..1em}

\begin{document}
\let\WriteBookmarks\relax
\def\floatpagepagefraction{1}
\def\textpagefraction{.001}
\shorttitle{}
\shortauthors{}

\title [mode = title]{Short-term Maintenance Planning of Autonomous Trucks for Minimizing Economic Risk}

\author[1]{Xin Tao}
\cormark[1]
\ead{taoxin@kth.se}

\author[1,2]{Jonas Mårtensson}
\ead{jonas1@kth.se}

\author[3]{Håkan Warnquist}
\ead{hakan.warnquist@scania.com}

\author[1]{Anna Pernestål}
\ead{annapern@kth.se}

\address[1]{Integrated Transport Research Lab, KTH Royal Institute of Technology, SE-100 44  Stockholm, Sweden.}
\address[2]{Division of Decision and Control Systems, KTH Royal Institute of Technology, SE-100 44 Stockholm, Sweden.}
\address[3]{Scania CV AB, SE-151 87 Södertälje, Sweden.}
\cortext[cor1]{Corresponding author. }

\begin{abstract}
New autonomous driving technologies are emerging every day and some of them have been commercially applied in the real world. While benefiting from these technologies, autonomous trucks are facing new challenges in short-term maintenance planning, which directly influences the truck operator's profit. In this paper, we implement a vehicle health management system by addressing the maintenance planning issues of autonomous trucks on a transport mission. We also present a maintenance planning model using a risk-based decision-making method, which identifies the maintenance decision with minimal economic risk of the truck company. Both availability losses and maintenance costs are considered when evaluating the economic risk. We demonstrate the proposed model by numerical experiments illustrating real-world scenarios. In the experiments, compared to three baseline methods, the expected economic risk of the proposed method is reduced by up to $47\%$. We also conduct sensitivity analyses of different model parameters. The analyses show that the economic risk significantly decreases when the estimation accuracy of remaining useful life, the maximal allowed time of delivery delay before order cancellation, or the number of workshops increases. The experiment results contribute to identifying future research and development attentions of autonomous trucks from an economic perspective.
\end{abstract}


\maketitle

\section{Introduction}
\subsection{Maintenance planning of autonomous trucks}

\noindent \begin{table*}\rmfamily 
\normalsize
\noindent\begin{tabular} {|p{0.8 cm}  p{6.7 cm}  p{1.0 cm} p{6.7 cm} |}
\hline 
& & & \\
\large Acronyms & & $c_{m|nb}$  & workshop maintenance cost of a truck that does  \\
& & & not break down\\
AV & Autonomous Vehicle & $c_{tow}$ & tow truck service fee   \\
RUL & Remaining Useful Life & $c_{f}$  & fixed cost of the tow truck service  \\
MER & Minimal Economic Risk  & $c_{var}$  & variable cost factor of the tow truck service  \\
IVHM & Integrated Vehicle Health Management & $d_{tow}$ & driving distance of the tow truck  \\
& &$u_j(l_j)$ & economic utility function of impact $j$   \\
\large  Symbols & & $L_{i}$ & economic loss given decision $i$   \\
&  &  $L_{j|i}$ &  economic loss caused by impact $j$ given decision\\
$i$ & maintenance decision $i$, $i \in \{ {wr,wn,cn}\}$ & & $i$  \\
$j$ & decision impact $j$, $j \in \{ {al,mc}\}$  & $l_{j|i}$ & value of decision impact $j$ given decision $i$  \\
$t_{m|nb}$ & workshop maintenance time of a truck that does  & $f_i(t)$ & probability density function of RUL given  \\
&  not break down & & decision $i$ \\
$t_{m|b}$ & workshop maintenance time of a truck that  & $F_i(t)$ & cumulative distribution function of RUL given   \\
& breaks down & & decision $i$ \\
$v_i$ & speed of the truck driving to the workshop  & $\alpha_i$ & shape parameter of $f_i(t)$ \\
$v_l$ & speed of a loaded tow truck    &  $\beta_i$ & scale parameter of $f_i(t)$ \\
$v_{ul}$ & speed of an unloaded tow truck  &  ${p_i}(nb,t)$ & probability density that the truck does not\\
${d_{tw}(t)} $ & distance from the truck to the nearest available & & break down at time $t$ given decision $i$ \\ 
& workshop if the truck continues to drive to the   &  $t_{min}$ & maximal allowed time of delivery delay without \\
& customer and breaks down at time $t$ & & penalty \\
${d_{wc}(t)}$ & distance from the nearest available workshop to  & $t_{max}$ & maximal allowed time of delivery delay before\\
 & the customer if the truck continues to drive to    & & order  cancellation\\
& the customer and breaks down at time $t$ & $pe_{max}$ & penalty of order cancellation \\  
$t_{w|i}$ & the time for the truck to reach the workshop &  $d_a$ & the distance from the alarm location to the \\  
& without breakdown given decision $i$ &  & highway entrance   \\
$l_{j|i,b(t)}$ & value of decision impact $j$ if the truck takes  & $\mathrm{E}[L_{i|d_a}]$ & economic risk given decision $i$ and $d_a$\\
& decision $i$ and breaks down at time $t$ & $bm_i$ & baseline method $i$   \\
$l_{j|i,nb}$ & value of decision impact $j$ if the truck takes & ${E\!E\!R}(bm_i)$ & expected economic risk of baseline method $i$  \\ 
& decision $i$ and does not break down &${E\!E\!R}(pm)$ & expected economic risk of the proposed method \\ 
$v_n$ & normal speed of the truck & $R_i$ &  the reduction ratio of the expected economic risk  \\
${d_{c}}$ & distance from the alarm location to the customer & & of the proposed method over that of $bm_i$   \\
$t_{sc}$ & time of scheduling a tow truck  & $\mathrm{Var}_i$ & variance of RUL given decision $i$\\
$c_{m|b}$ & workshop maintenance cost of a truck that & $\mathrm{E}_i$  &  expectation of RUL given decision $i$\\
& breaks down & & \\
\hline
\end{tabular}
\end{table*}

In recent years, research and development on Autonomous Vehicles (AV) have accelerated. Although much of the attention has focused on autonomous cars, there is consensus among industry experts that autonomous trucks are likely to become commercially available first \cite{crandall2016driverless}. Until now research on autonomous trucks has been limited, but there is a rapidly growing interest \cite{neuweiler2017autonomous}. More specifically, scientific research that deals with the maintenance of autonomous trucks is very sparse, or even lacking \cite{vskerlivc2020analysis}. However, for the practicability of autonomous trucks in the real world, it is critical to facilitate them with effective maintenance planning schemes.

Increased maintenance performance can bring in growing profitability to the industry \cite{FACCIO201485}\cite{cachada2018maintenance}. Maintenance causes monetary cost as well as availability loss. While monetary cost is a direct economic loss, availability loss causes economic loss indirectly. Increased maintenance performance leads to high availability, thus increasing short-term profit. Furthermore, high availability also increases long-term profit by increased trust and order demands from customers \cite{bernspaang2011measuring}. Therefore, making maintenance plans that balance availability and maintenance cost is important for the overall profit of autonomous trucks \cite{biteus2017planning}.

Advanced AV technologies contribute to maintenance planning in various ways. Firstly, the development of sensing technology and intelligent algorithms enable more advanced fault diagnosis and prognosis technologies. As a result, it is possible to obtain more accurate fault information before the fault evolves into a failure. Meanwhile, the development of communication technologies and the application of the Internet of Things to AV enable the interconnection of vehicles, infrastructures, and information resources, which provides more maintenance alternatives and information for maintenance planning. 

Besides the benefits, these technologies also bring new challenges, especially when the autonomous truck is on a freight transport mission and faults occur abruptly. In this situation, maintenance must be planned and executed in the short-term, which is challenging for several reasons:
\begin{itemize}[noitemsep]
  \item \textbf{Absence of an onboard decision-maker.}  On a manually driven vehicle, the driver also acts as a sensor, inspector, and decision-maker. When no driver is present, the vehicle needs to make a short-term maintenance plan by itself.
  \item \textbf{Increase in continuous operating time.} Without the constraints of working hours of the driver, the autonomous truck is expected to drive continuously for long hours (close to 24/7 operation) \cite{Engholm2020}. This increased utilization, as an advantage of AV, reduces room for planned maintenance.  As a result, more unplanned maintenance is likely to happen, limiting the utilization of the vehicle.
  \item \textbf{Increase in complexity and uncertainty of faults.} With more complex AV technologies, more potential abnormal conditions can occur and cause an element or component to fail, referred to as faults. As a result, the complexity and uncertainty of faults increases. 
\end{itemize}

\subsection{Problem and contribution statement}
In this paper, we address the challenges of short-term maintenance planning for autonomous trucks using a risk-based decision-making method. This method enables the integrated usage of various types of information, including dynamically changing information such as the Remaining Useful Life (RUL) of the faulty component, truck location, etc. When a fault is detected on an autonomous truck during a transport mission, the proposed model provides an optimal maintenance decision with Minimal Economic Risk (MER). 

The main contributions of this paper include: 
\begin{itemize}[noitemsep]
    \item identifying the information and information flow required for maintenance planning in the context of autonomous trucks on a freight transport mission; 
    \item developing a maintenance planning model of autonomous trucks using a risk-based decision-making method. This model identifies the maintenance plan with MER considering the economic risk caused by both availability loss and maintenance costs;
    \item performing sensitivity analyses of various parameters in the maintenance planning model, thus bringing in ideas on research and development attention of the maintenance planning of autonomous trucks. 
\end{itemize}

The remainder of the paper is organized as follows: In Section \ref{sec2}, a brief literature review is presented. Section \ref{sec3} describes a system design of an Integrated Vehicle Health Management (IVHM) system and an overall scheme of maintenance planning. In Section \ref{sec4}, a maintenance planning model is constructed, including the selection of decision alternatives, models of the decision impacts, including availability loss and maintenance cost and a risk-based decision-making method. In Section \ref{sec5}, numerical experiments are conducted, together with sensitivity analysis of model parameters and a discussion on practical issues. Section \ref{sec6} makes a conclusion and offers some suggestions for future work. 
\section{Literature review} \label{sec2}
Maintenance planning is an established research area with a lot of engineering applications, especially in complex industrial systems like infrastructure \cite{gong2020condition}, manufacturing \cite{lundgren2018quantifying}, transport \cite{tavares2016vehicles}\cite{lin2019optimization}, and electricity \cite{froger2016maintenance}. It is considered as one of the main outcomes of a decision-making module of a prognosis and health management system \cite{atamuradov2017prognostics} \cite{li2020systematic}. There is also considerable research on vehicle maintenance. It was highlighted in \cite{ezhilarasu2019application} \cite{zhang2015integrated} that it is important to utilize the capabilities of Integrated Vehicle Health Management to enable effective and efficient maintenance and operation of the vehicle. One of these capabilities is remote diagnosis and maintenance, which has drawn attention in the automotive industry \cite{shafi2018vehicle} \cite{tavares2016vehicles}. In \cite{bouvard2011condition}, an optimization of grouping maintenance operations for a commercial heavy vehicle was addressed to reduce the global maintenance cost of the system. In \cite{8790142}, a vehicle fleet maintenance scheduling optimization problem was solved by a multi-objective evolutionary algorithm. Despite these efforts on vehicle maintenance, they did not consider short-term and unplanned maintenance and the risk of failure during a transport mission. In this paper, we consider these issues, especially in the context of autonomous trucks.

Research communities have realized the importance and identified the challenges of vehicle maintenance planning, especially with the development of AV technologies. A recent literature review in the area of maintenance management performance demonstrates the need for detailed quantitative analyses to properly choose the maintenance strategy \cite{nowakowski2018evolution}.  Moreover, analyses in \cite{vskerlivc2020analysis} show that scientific research that deals with truck maintenance, especially modern truck maintenance guidelines are lacking. With the development of AV technologies, new demands and research topics for fault handling and maintenance planning are emerging. In \cite{tavares2016vehicles}, the necessity to study the use of emerging technologies, especially data processing and communications technologies in vehicle maintenance is identified. According to \cite{wadud2017fully}, the effects of full automation on maintenance and repair are not well understood. With these research demands and challenges identified, it is important to tackle and implement them by proposing effective solutions. 

There are several typical and well-adopted maintenance strategies, including reactive maintenance, planned maintenance, predictive maintenance, and proactive maintenance \cite{nowakowski2018evolution}  \cite{tavares2016vehicles}. Among them, predictive maintenance uses non-destructive sensors and prognosis techniques to identify the presence of faults and failures and estimating the RUL of the system \cite{tavares2016vehicles}. It is used for infrequent failures when the repair is extremely costly and becomes the common practice in the vehicle industry\cite{shafi2018vehicle}. This maintenance strategy have attracted considerate attention in the recent years, with the development of prognosis technologies \cite{bouvard2011condition} \cite{bousdekis2015proactive} \cite{ellefsen2019comprehensive}. As an important information resource of maintenance planning, prognosis technologies are developing  rapidly supporting by various machine learning methods \cite{le2016remaining} \cite{chen2019hidden} \cite{liao2018uncertainty}. Useful as it is, this strategy may not be enough for short-term maintenance planning of autonomous trucks on a freight transport mission, because the truck company considers the overall profit of the delivery task rather than the risk of truck failure only. 

Depending on the time span, maintenance planning can be classified into long term and short term. Long-term planning is commonly considered, especially the life cycle of the system \cite{gong2020condition} \cite{bernspaang2011measuring} \cite{frangopol2017bridge}. There were also researches on short-term maintenance planning \cite{yang2017novel} \cite{sadeghian2019risk}, which were less up-to-date and have attracted much less attention than long-term ones. However, in the context of maintenance planning for autonomous truck, the need for short-term maintenance may increase as a result of more intensive and non-stop driving \cite{Engholm2020}. 

Researchers have applied various theories and methods to solve maintenance planning problems, such as decision theory, optimization methods, and risk assessment. In \cite{sanchez2009addressing}, a multi-objective optimization problem was formulated for maintenance planning based on unavailability and cost criteria. An intelligent decision making method, particle swarm optimization was utilized in \cite{wu2018optimizing} to solve the high-level maintenance planning problem of the electric multiple unit train. Deep reinforcement learning is used in \cite{zhang2020deep} for condition-based maintenance planning of multi-component systems. Among a rich variety of methods, risk-based maintenance methodology provides a tool for maintenance planning and decision making to reduce the probability of failure of equipment and the consequences of failure \cite{gong2020condition}\cite{sadeghian2019risk}\cite{yang2017novel}. In the context of autonomous trucks on a freight transport mission, applying these methods to tackle the new challenges is important. Among them, risk-based decision-making is well suitable for tackling issues with dynamic and uncertain nature.
\section{System design and overall scheme} \label{sec3}
\subsection{System design} \label{sec3.1}
\begin{figure}[ht]
\centering
\includegraphics[width= .6\linewidth]{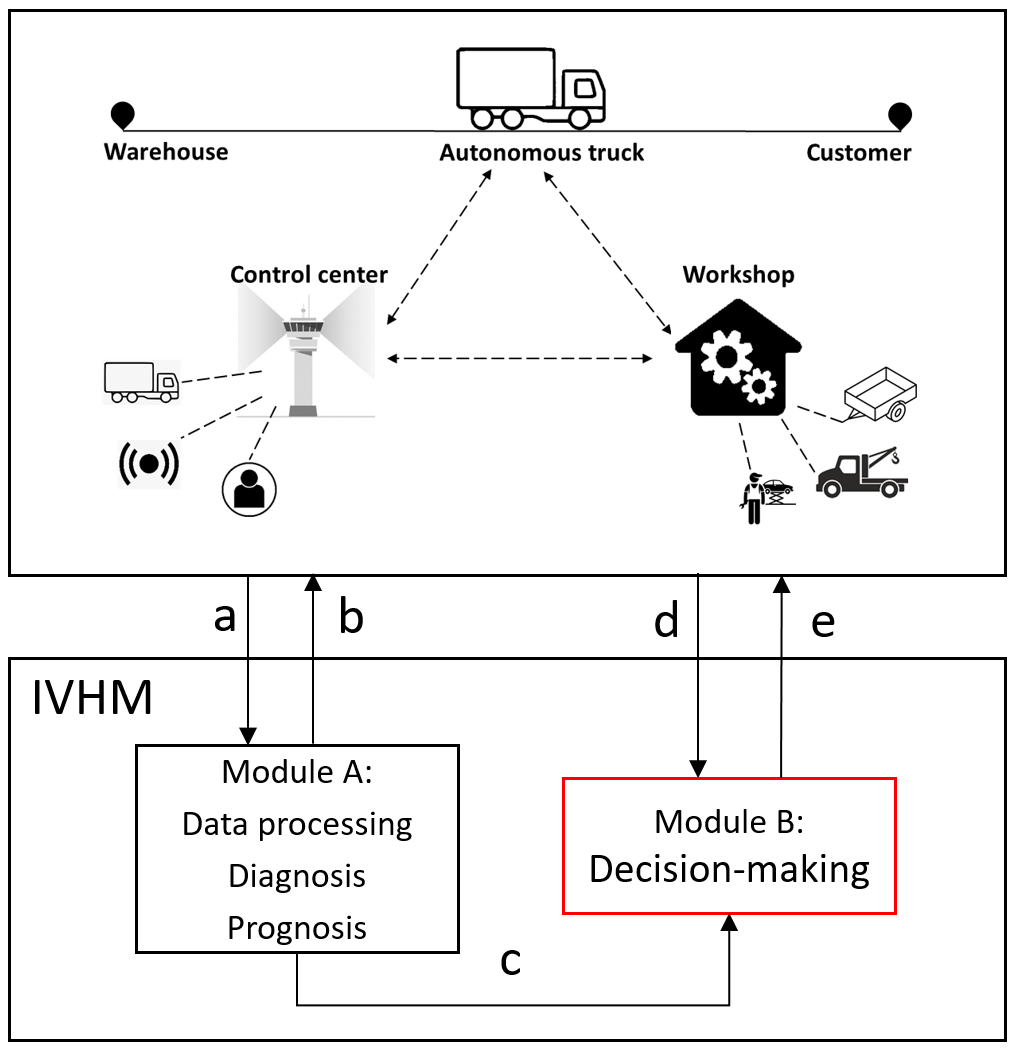}
\caption{An IVHM system design adapted to autonomous trucks.}
\label{fig1}
\end{figure}

In \cite{zhang2015integrated} \cite{ezhilarasu2019application}, a generic architecture of an IVHM system was proposed, which described different modules in the system and their connections. Based on this architecture, we construct an IVHM system design adapted to autonomous trucks, see Fig. \ref{fig1}. This system design enables information flows between different system actors and IVHM modules. Specifically, there are two main parts, shown as the two large boxes in Fig. \ref{fig1}. 

The upper box depicts the system actors involved in the delivery and maintenance planning tasks. The lower box presents the essential modules of the IVHM system, including a data processing module, a diagnosis module, a prognosis module, and a decision-making module. The decision-making module is emphasized as Module B, which is the focus of this paper. Other modules are here lumped together and referred to as Module A.

In this system design, different system actors, Module A, and Module B are interconnected with information flows, shown as solid single-headed arrows in Fig. \ref{fig1}. In total, there are five information flows:
\begin{itemize}[noitemsep]
\item Information flow \textbf{a} is provided by system actors and used for fault diagnosis and prognosis. This information includes sensor data, historical fault data, etc.; 
\item Information flow \textbf{b} is the fault information, such as fault type and fault location. It is input to system actors for further usages, such as estimating the maintenance time and cost.
\item Information flow \textbf{c} is the fault information used for decision-making, such as the RUL and its distribution;
\item Information flow \textbf{d} is provided by system actors and used for decision-making. This information includes fault alarm location, workshop information, etc.
\item Information flow \textbf{e} is about the execution of maintenance decisions, such as the destination and driving speed.
\end{itemize}

The decision from the IVHM system (information flow e) includes two types of actions, including operational actions, such as maintenance interventions, hardware/software reconfigurations, and fault-tolerant control, and design-based actions, such as adding and/or replacing sensors and redesigning component placement \cite{atamuradov2017prognostics}. In this paper, maintenance intervention is considered, which is a typical operational decision.

The system actors include the truck, the warehouse, the customer, the control center, and the workshop. While delivering goods from the warehouse to the customer, the autonomous truck is connected to the control center and the workshop wirelessly. The control center can monitor, communicate, and schedule remotely, while the workshop owns maintenance resources like personnel, towing service, and trailer. Wireless communication enables these actors to connect remotely and exchange information in real-time to facilitate maintenance planning, shown as the dashed double-headed arrows in Fig. \ref{fig1}. The modules in the IVHM system are either centrally installed in the autonomous truck or distributively installed in the truck and the control center.

\subsection{Overall scheme of maintenance planning} \label{sec3.2}
\begin{figure} [ht]
\centering
\includegraphics[width= .6 \linewidth]{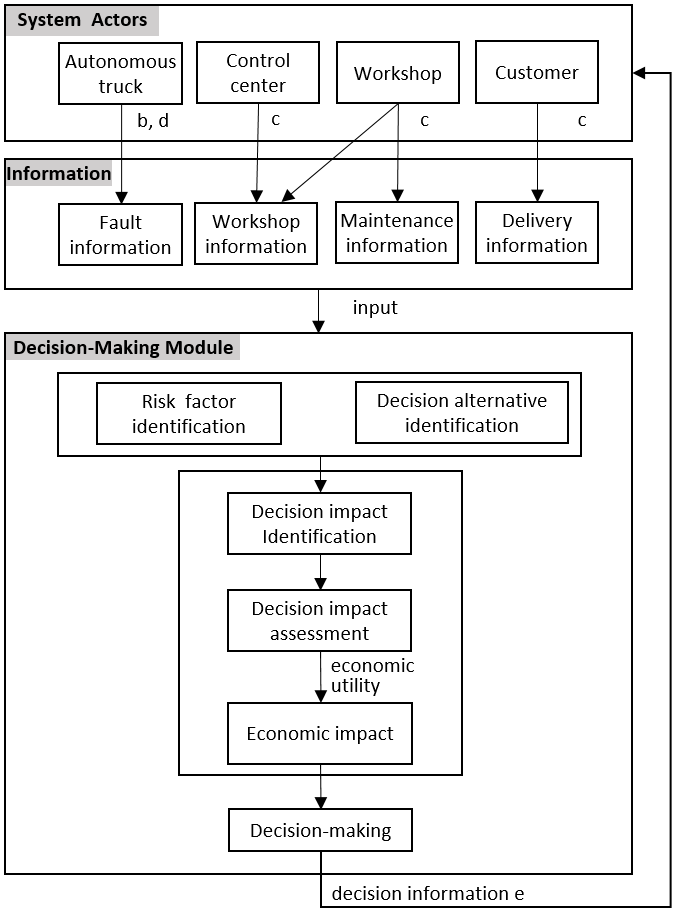}
\caption{Overall scheme of maintenance planning.}
\label{fig2}
\end{figure}

In this subsection, we propose an overall scheme of maintenance planning, see Fig. \ref{fig2}. This scheme corresponds to the decision-making module (Module B) in Fig. \ref{fig1} 

As shown in Fig. \ref{fig2}, the decision-making module obtains various information inputs from different system actors to facilitate maintenance planning. These information inputs correspond to the information flow c and d in Fig. \ref{fig1}. Specifically, there are mainly four types of information input, including:
\begin{itemize}[noitemsep]
\item \textbf{fault information} provided by the truck. This includes both the fault diagnosis and prognosis information provided by Module A on the truck (information flow c) and other fault information (a subset of information flow d), such as the fault alarm location.
\item \textbf{workshop information} provided by the control center or the workshops (a subset of information flow d), such as the workshop locations, availability, and capability. 
\item \textbf{maintenance information} provided by the workshop (a subset of information flow d), such as the maintenance time, maintenance cost, and tow truck service fee.
\item \textbf{delivery information} provided by the customer (a subset of information flow d), such as the freight value and delay penalty. 
\end{itemize}

We assume that both the workshops or the control center can provide the workshop information to the truck upon request. In this paper, we assume that the control center has a centralized information pool that stores history information and collects real-time information from different workshops. Therefore, the truck obtains the workshop information from the control center instantly without any delay.

Given the information inputs, the decision-making module is constructed, shown as the bottom box in Fig. \ref{fig2}. Firstly, the decision alternatives and risk factors are identified and specified. If there is a human driver, the driver could perform these tasks. But for an autonomous truck, the decision-making module needs to perform these tasks automatically. Note that the automation of these tasks is not the scope of this paper. Therefore, we predefine the decision alternatives and risk factors manually, which are presented in Section \ref{sec4.1}.

In this subsection, we identify the potential impacts caused by executing a maintenance decision, such as safety, availability, environment, economy and so on. By analysing the significance and relevance of the decision impacts to economic risk, we identify the decision impacts to be assessed quantitatively and perform the assessment considering risk factors.

Finally, a maintenance decision is made by selecting the decision with MER, which is sent to the system actors for execution.
\section{Maintenance planning model} \label{sec4}
In this section, we further implement the decision-making module in Fig. \ref{fig2}. First, we select the decision alternatives and analyze the risk factors. Then we present a risk-based decision-making model, which follows a typical probabilistic risk assessment method \cite{aven2013uncertainty}. 

\subsection{Risk factors and decision alternatives} \label{sec4.1}

In this paper, we consider component degradation faults. When a component degrades over time, there is a risk that the fault causes component failure, so that the ability of the component to perform its function as required is terminated. If the failure happens, we assume that the truck needs to stop driving and wait for towing service and maintenance service, an event that can cause high economic loss. In order to make a maintenance plan that reduces economic risk, we need information about the risk of component failure over time.

\begin{figure}[ht]
\centering
\includegraphics[width= .6\linewidth]{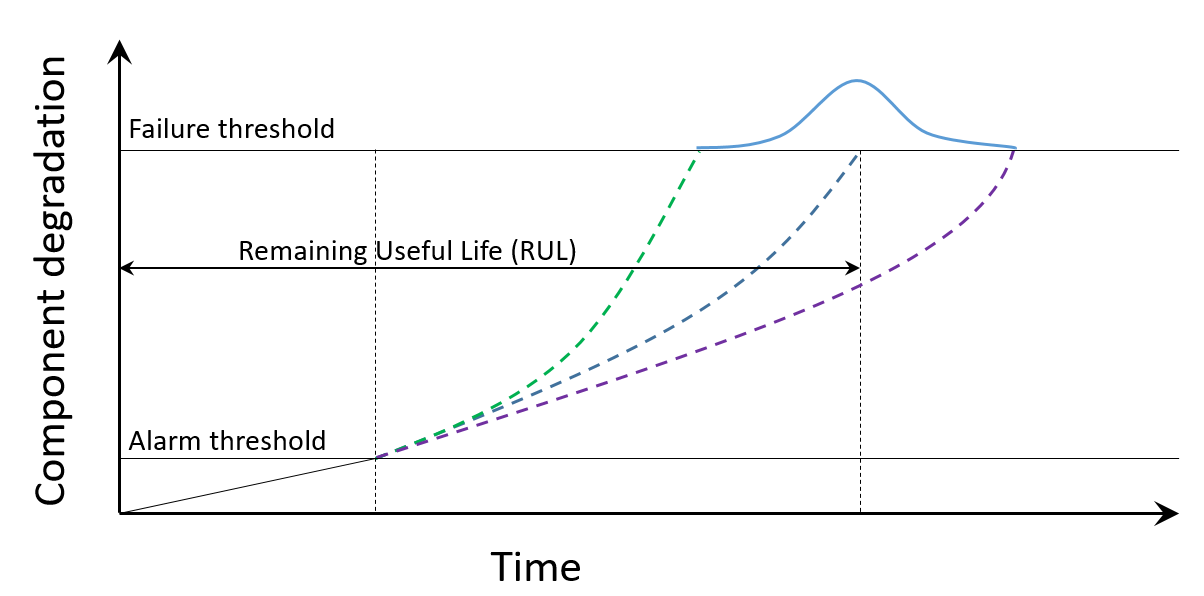}
\caption{Degradation process of a faulty component overtime.}
\label{fig:3}
\end{figure}

The risk of component failure over time is predicted depending on the component degradation process. This process not only depends on component properties but also on other factors, like the operating environment, operating intensity, and so on. In Fig. \ref{fig:3}, the degradation process of a component is schematically depicted \cite{bouvard2011condition}. When the degradation evolves and reaches a predefined alarm threshold, a fault alarm is on and the prognosis module starts to predict the time when a failure occurs, also known as RUL. The dashed lines in Fig. \ref{fig:3} represent different possible degradation processes and the blue curve represents a probabilistic distribution of the estimated RUL.

Note that the RUL distribution depends on the prediction method and does not necessarily follow a specific type of distribution. The prediction of RUL involves a lot of work, including obtaining fault data, building the prediction model, assessing uncertainties, and so on. There are massive existing and ongoing researches on RUL prediction, many of which are based on machine learning techniques. In this paper, we take the RUL distribution as the input information from the prognosis module. 

Having obtained the risk of component failure, represented by the RUL distribution, we predefine the decision alternatives as discussed in Section \ref{sec3.2}. First, we derive two potential ways to reduce the risk of component failure, including being maintained as soon as possible and decreasing the degradation speed of the fault. The former way can be implemented by the action of driving to the nearest workshop, while we assume the latter way can be implemented by the action of reducing the driving speed. Note that some degradation faults are dependent on driving speed while others are not. Therefore, this assumption is an example formulation and might not be valid for all types of faults.

In the end, we define three decision alternatives depending on whether taking the actions and considering the industry practice: 

\begin{itemize}[noitemsep]
\item Decision $i=wr$: Drive to the nearest available workshop at a reduced speed, do the maintenance, and continue the delivery task.
\item Decision $i=wn$: Drive to the nearest available workshop at a normal speed, do the maintenance, and continue the delivery task.
\item Decision $i=cn$: Drive to the customer at a normal speed. After the delivery task, go to the nearest workshop for maintenance.
\end{itemize}

The notation of different decisions consists of two letters. The first letter indicates the first destination of the truck, with $w$ for the workshop and $c$ for the customer. The second letter indicates the driving speed of the truck before reaching the workshop, with $r$ for a reduced speed and $n$ for a normal speed.

The three decision alternatives represent three risk attitudes of the decision-maker towards component failure. Specifically, decision $wr$ is a risk-avoidance type, which tries to reduce the risk of component failure by taking both two actions; decision $wn$ is a risk-neutral type, which tries to reduce the risk of component failure taking one action, i.e. driving to the workshop; decision $cn$ is a risk-seeking type, which takes no action, ignores the risk of component failure and continues the task.

Note that these three decision alternatives do not constitute an exhaustive collection of all possible decisions. We use such simplification of decision alternatives to increase the understanding of the problem and draw more general conclusions.

\begin{figure} [ht]
\centering
\includegraphics[width= .6 \linewidth]{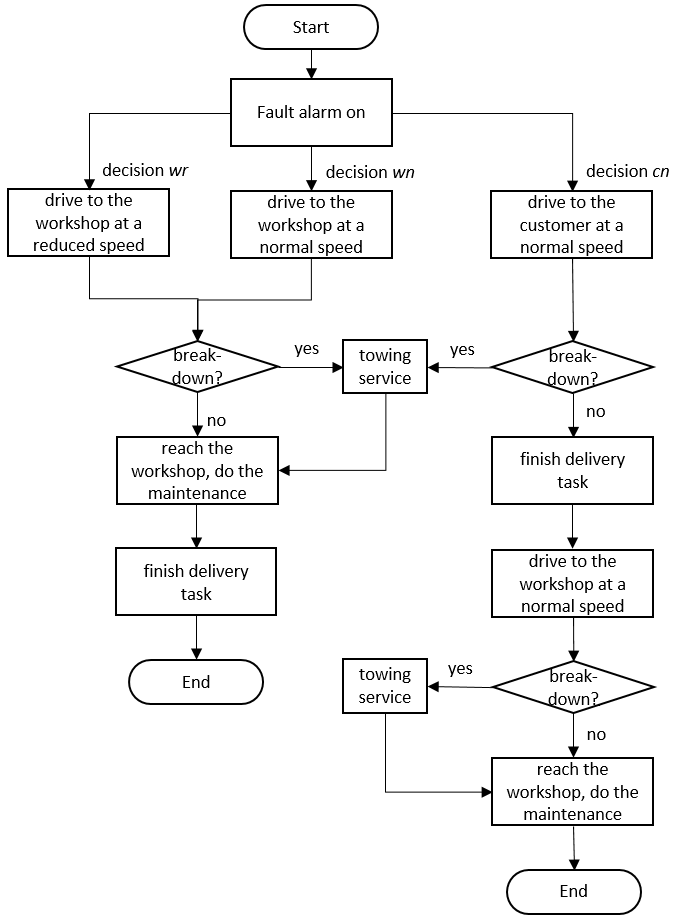}
\caption{Execution process of different maintenance decisions,  starting with noticing a fault alarm and ending when two tasks (the delivery task and the maintenance task) are finished.}
\label{fig4}
\end{figure}

Assume that if the truck breaks down, a tow truck is scheduled by the nearest available workshop. After maintenance in the workshop, the truck continues the delivery task, during which the risk of truck breakdown is negligible.  The executing process of the three proposed decision alternatives is presented in a flowchart in Fig. \ref{fig4}. This flowchart starts with noticing that a fault alarm is on and ends when two tasks (the delivery task and the maintenance task) are finished.

\subsection{Decision impacts} \label{sec4.2}
In this subsection, we analyze the potential impacts caused by executing a maintenance decision, identify the decision impacts relevant to economic risk, and assess the relevant impacts quantitatively. 

\subsubsection{Decision impact identification }\label{sec4.2.1}
When the fault alarm is on, executing a maintenance decision can cause various types of decision impacts, mainly including availability loss, maintenance cost and safety. Among them, safety impact is usually assessed by other safety controllers instead of the maintenance planner, although the actions of safety controllers can influence the executability of a maintenance plan. There are massive ongoing research on developing safety controller for autonomous vehicles \cite{cui2019review} \cite{machin2016smof}. In this paper, we do not consider safety impact and focus on impacts that are directly  relevant to economic risk. While maintenance cost is a direct economic loss, availability loss can cause indirect economic loss in several ways, such as delivery delay and order cancellation. For freight types with special requirements like refrigeration, availability loss may cause damage on the goods. In practice, if these situations happen, an economic penalty of the truck operator is usually required as compensation to the customer. The penalty amount is usually part of the contract or agreement between the truck company and the customer when the delivery task is specified. This penalty is directly related to the economic impact of availability loss. As a result, two decision impacts, i.e. availability loss  and maintenance cost are considered in this paper. An overall consideration of cost and availability is commonly adopted in maintenance planning \cite{durazo2018autonomous}\cite{ERKOYUNCU201753}. 

The decision impact is denoted as $j \in \{al, mc\}$, where $al$ is availability loss and $mc$ is maintenance cost. To evaluate the decision impacts quantitatively, we need to model decision impacts with more details. The decision impact models can be different depending on available maintenance resources and maintenance details etc. In the following two subsections, we present two decision impact models as parts of the maintenance planning model. Note that these are not the only ways to model them. Instead, we aim to illustrate the complexity of this modeling process due to the mobility of the autonomous truck.

\subsubsection{Availability loss} \label{sec4.2.2}
We define the availability loss as the difference between the planned delivery time and the actual delivery time. The actual delivery time consists of two types: 
 \begin{itemize}[noitemsep]
 \item \textbf{Maintenance time}. Maintenance time includes the time of scheduling personnel, preparing spare parts, and repairing the truck, etc \cite{camci2009system}. Since the modeling of maintenance activities is not the main focus of this paper, a total maintenance time is used. However, the influence of the truck state on maintenance time is considered. If the truck breaks down before maintenance, the time of fixing the truck can be longer than fixing a faulty component. For example, the failure of gear may damage the whole transmission system, and the time of fixing the transmission system is much longer than the time of replacing a gear. 
 \item \textbf{Travel time}. Travel time refers to the time spent on the road to the workshop and to the customer. If the truck does not break down on the way to the workshop, the travel is fulfilled by the truck itself. If it does, the travel time also includes the time of scheduling a tow truck and the time the tow truck spends on the road.
\end{itemize} 

The maintenance time of a truck that breaks down, $t_{m|b}$, and that if it does not break down, $t_{m|nb}$, are set to constants, where $t_{m|b} \geq t_{m|nb}$. The travel time depends on the driving speed of the truck to the workshop $v_i$ given decision $i$. If the truck breaks down, the tow truck speed also influences the travel time. For safety reasons, the speed of a loaded tow truck $v_l$ is restricted and slower than that of an unloaded tow truck $v_{ul}$.

Set the time when the fault alarm is on as $t=0$. If the truck continues to drive to the customer and breaks down at time $t$, the control center provides the truck with the distance from the truck to the nearest available workshop $d_{tw}(t)$, and the distance from this workshop to the customer $d_{wc}(t)$. The distance from the alarm location to the nearest available workshop is $d_{tw}(0)$, and the distance from that workshop to the customer is $d_{wc}(0)$.

For decision $i\in \{ wr, wn\}$, if the truck does not break down on the road to the workshop, the time of driving from the alarm location to the nearest available workshop is
\begin{equation}
    t_{w|i} = {d_{tw}(0)}/v_i.
\end{equation}

The actual delivery time consists of the time spent on three events:\\
1) truck driving to the workshop;\\
2) maintenance in the workshop; \\
3) truck driving from the workshop to the customer.

The value of decision impacts depends on if the truck breaks down before reaching the workshop. When executing decision $i$. If the truck takes decision $i$ and breaks down at time $t\in [0,{t_{w|i}}]$, the value of impact $j$ is denoted as  $l_{j|i,b(t)}$. If it does not break down, the value of impact $j$ as $l_{j|i,nb}$. 

If the breakdown does not occur, the total availability loss is 
\begin{equation}
    {l_{al|i, nb}} = t_{w|i} + t_{m|nb} + d_w^c(0)/v_n - d_c/v_n,
\end{equation}
where $v_n$ is the normal speed of the truck and $d_c$ is the distance from the alarm location to the customer. 

If the truck breaks down at time $t$, $t \in [0,t_{w|i}]$, the actual delivery time consists of the time spent on six events: \\
1) truck driving to the workshop;\\
2) scheduling a tow truck;\\
3) tow truck driving to the truck;\\
4) tow truck towing the truck to the workshop;\\
5) maintenance in the workshop; \\
6) truck driving from the workshop to the customer.

The total availability loss is
\begin{equation} \label{eq:3}
        {l_{al|i,b(t)}} 
        = t +{t_{sc}} + (d_{tw}(0)-v_it)/{v_{ul}}   
         + (d_{tw}(0)-v_it)/{v_l} + t_{m|b} 
         + d_{wc}(0)/v_n - d_c/v_n, 
\end{equation}
where $t_{sc}$ is the time of scheduling a tow truck.

For decion $i=cn$, if the truck does not break down on the road to the workshop, there is no availability loss, i.e.
\begin{equation}
    {l_{al|cn,nb}} = 0.
\end{equation}

The time of driving to the customer, $t_{c|{cn}}$ is 
\begin{equation}
    t_{c|{cn}}= d_c/v_{cn}.
\end{equation}

The time for the truck driving to the workshop is
\begin{equation} \label{eq:6}
    {t_{w|{cn}}} = t_{c|{cn}} + d_{wc}(t_{c|{cn}})/v_{cn}.
\end{equation}

If the truck breaks down after the delivery task, it still takes time to fix the truck. However, the availability of the truck for the current delivery task is not affected. Therefore, for $t \in [t_{c|{cn}},t_{w|{cn}}]$, 
\begin{equation}
    l_{al|cn, b(t)} = 0.
\end{equation}

If the truck breaks down at time $t$ on the way to the customer, $t \in [0,{t_{c|{cn}}}]$, the actual delivery time consists of the time spent on six events, which are the same as that of decision $wr$ and $wn$ listed above. The total availability loss is
\begin{equation}
        {l_{al|{cn},b(t)}}
         = t +{t_{sc}} + d_{tw}(t)/{v_{ul}} + d_{tw}(t)/{v_l} 
          + t_{m|b}+ d_{wc}(t)/v_n - t_{c|{cn}}. 
\end{equation}

\subsubsection{Maintenance cost} \label{sec4.2.3}
We define the maintenance cost as the direct monetary cost by taking a maintenance decision. There are mainly two types of maintenance costs: 
\begin{itemize}[noitemsep]
    \item \textbf{Maintenance cost in the workshop}. The maintenance cost in the workshop for repairing the truck can be further specified as personnel cost and costs of spare parts etc. Since this is not the main focus of this paper, a total maintenance cost is used. However, the influence of the truck state on workshop maintenance cost is considered. If the truck breaks down before reaching the workshop, the workshop maintenance cost can be higher than the cost if it does not. For example, a brake failure may cause a collision and damage the truck, resulting in much more maintenance cost than fixing a brake with faults. 
    \item \textbf{Tow truck service fee}. If the truck breaks down on the road, a tow truck is scheduled, which generates a service fee. 
\end{itemize} 

The workshop maintenance cost of a truck that breaks down $c_{m|b}$ and that if it does not break down $c_{m|nb}$ are set as constants, where $c_{m|b}\geq c_{m|nb}$.

For all the maintenance decisions, if the truck does not break down before reaching the workshop, there is no tow truck service but only workshop maintenance cost, i.e.
\begin{equation}\label{eq:9}
    {l_{mc|i, nb}} = c_{m|nb}.
\end{equation}

If the truck breaks down before reaching the workshop, there is a tow truck service fee. We model the tow truck service fee as the sum of a fixed cost and a cost linearly related to the distance the tow truck drives, denoted as $c_{tow}$ and expressed as:
\begin{equation}\label{eq:10}
    c_{tow} = {c_{f}} + c_{var}d_{tow},
\end{equation}
where ${c_{f}}$ is the fixed tow truck service fee and ${c_{var}}$ is the variable cost factor for every kilometer the tow truck drives. $d_{tow}$ is the driving distance of the tow truck. 

For decision $i\in \{ wr,wn\}$, if the truck breaks down at time $t$, $t \in [0,t_{w|i}]$, the distance the tow truck drives is $2(d_{tw}(0)-v_it)$. The maintenance cost $l_{mc|i,b(t)}$ is the sum of the tow truck service fee and the maintenance cost:
\begin{equation} \label{eq:11}
    {l_{mc|i, b(t)}} = c_{m|b} + {c_{f}} + 2(d_{tw}(0)-v_it)c_{var}.
\end{equation}

For decision $i=cn$, if the truck fails at time $t$, $t \in [0,t_{w|3}]$, the distance the tow truck drives is $2d_{tw}(t)$. The maintenance cost $l_{mc|cn,b(t)}$ is the sum of the tow truck service fee and the workshop maintenance cost:
\begin{equation} \label{eq:12}
    {l_{mc|cn,b(t)}} = c_{m|b} + {c_{f}} + 2d_{tw}(t)c_{var}.
\end{equation}
\subsection{Risk-based decision-making} \label{sec4.3}
In this subsection, we implement the maintenance planning model using a risk-based decision-making method. Considering the risk of truck breakdown before reaching the workshop, the economic risk is assessed as the expected economic loss caused by availability loss and maintenance cost. To assess the decision impacts in an economic scale, we use economic utility functions of impact $j$ that converts decision impacts to economic impacts, denoted as $u_j(\bullet)$. As analyzed in section \ref{sec4.2.1}, maintenance cost is a direct economic loss, while availability loss can cause indirect economic loss, which can be derived from the agreement between the truck company and the customer. 

Denote the economic loss of taking decision $i$ as ${L_i}$, where $i\in \{wr,wn,cn\}$. The maintenance planning problem is formulated as an optimization problem of finding the maintenance decision that causes minimal expected economic loss:  
\begin{equation} \label{eq:13}
    i^*=\mathop {\arg }\limits_i {\rm{ }}\min (\mathrm{E}[{L_i}]).
\end{equation}

Economic risk $\mathrm{E}[{L_i}]$ is caused by decision impact $j$. Decision impacts $j$ can be either availability loss $al$ or maintenance cost $mc$, i.e. $j\in\{al, mc\}$. Economic risk $\mathrm{E}[{L_i}]$ can be expressed as
\begin{equation}\label{eq:14}
    \mathrm{E}[{L_i}] = \sum\limits_j{ \mathrm{E}[{L_{j|i}}]},
\end{equation}
where ${L_{j|i}}$ is the economic loss caused by impact $j$ given decision $i$.

Economic loss ${L_{j|i}}$ is the economic utility of the value of impact $j$ given decision $i$: 
\begin{equation} \label{eq:15}
    {L_{j|i}} = {u_j}({l_{j|i}}),
\end{equation}
where ${l_{j|i}}$ is the value of impact $j$ given decision $i$. 

Therefore, economic risk $\mathrm{E}[{L_{j|i}}]$ can be expressed as the expected economic utility of ${l_{j|i}}$:
\begin{equation}
    \mathrm{E}[{L_{j|i}}] = { \mathrm{E}[{u_j}({l_{j|i}})]}.
\end{equation}

To further model economic risk $\mathrm{E}[{L_{j|i}}]$ considering the risk of truck breakdown, the evolving process of a fault into a failure is first introduced. 

We assume in Section \ref{sec4.1} that a reduced driving speed slows down the degradation process, thus prolonging the RUL of the truck. With this assumption, the probability distributions of RUL are different for decisions with different driving speeds. Denote the probability density function and the cumulative distribution function of RUL given decision $i$ as $f_i(t)$ and $F_i(t)$ respectively.

Before the truck reaches the workshop, there is a risk that the truck breaks down on the road, which influence both availability cost and maintenance cost. We assume that the risk of truck breakdown after maintenance is negligible. Denote the time for the truck to reach the workshop without breakdown given decision $i$ as $t_{w|i}$. If a breakdown event $b$ occurs at time $t\in [0,{t_{w|i}}]$, the value of impact $j$, $l_{j|{i,b(t)}}$, is different from that if no breakdown occurs $l_{j|{i,nb}}$, i.e. $l_{j|{i,b(t)}} \ne l_{j|{i,nb}}$. The economic risk $\mathrm{E}[{L_{j|i}}]$ can thus be written as: 
\begin{equation} \label{eq:17}
        \mathrm{E}[{L_{j|i}}] = \int_0^{t_{w|i}} {{u_j}({l_{j|{i,b(t)}}})f_{i}(t)}dt+
        \int_0^{t_{w|i}} {{u_j}({l_{j|{i,nb}}})p_{i}(nb,t)dt},
\end{equation}
where ${p_i}(nb,t)$ is the probability density that the truck does not break down at time $t$ given decision $i$.

If the truck does not break down on the way to the workshop, neither availability loss nor maintenance cost is influenced by the time of breakdown. Therefore, the latter part of Eq. (\ref{eq:17}) can be converted into:
\begin{equation}\label{eq:18}
    \int_0^{t_{w|i}} {{u_j}({l_{j|{i,nb}}})p_{i}(nb,t)dt}= {u_j}({l_{j|{i,nb}}})\int_0^{t_{w|i}} p_{i}(nb,t)dt,
\end{equation}
where
\begin{equation} \label{eq:19}
    \int_0^{t_{w|i}}p_{i}(nb,t)dt  = 1- \int_0^{t_{w|i}}f_{i}(t)dt.
\end{equation}

The integration of the probability density function $f_i(t)$ can be expressed with the cumulative distribution function $F(t)$ as:
\begin{equation}
    F(t) = \int {f(t)dt},
\end{equation}
which is used in this paper to simplify formula expressions and computation.

Therefore, Eq. (\ref{eq:19}) can be expressed as:
\begin{equation}\label{eq:21}
        \int_0^{t_{w|i}}p_{i}(nb,t)dt
        = 1 - (F_i(t_{w|i})-F_i(0))
        = 1- F_i(t_{w|i}),
\end{equation}
where $F_i(0) = 0$.

By first substituting Eq. (\ref{eq:21}) into Eq. (\ref{eq:18}), then substituting Eq. (\ref{eq:18}) into Eq. (\ref{eq:17}), we have that
\begin{equation} \label{eq:22}
        \mathrm{E}[{L_{j|i}}] = \int_0^{t_{w|i}} {{u_j}({l_{j|{i,b(t)}}})f_{i}(t)dt} +
        (1-F_i(t_{w|i})){u_j}({l_{j|{i,nb}}}).
\end{equation}

Then by first substituting Eq. (\ref{eq:22}) into Eq. (\ref{eq:14}), then substituting Eq. (\ref{eq:14}) into Eq. (\ref{eq:13}), the optimization problem in Eq. (\ref{eq:13}) is expressed as:
\begin{equation} \label{eq:23}
    i^*=    \mathop{\arg }\limits_i {\rm{ }}\min  (\sum \limits_j (\int_0^{t_{w|i}}{{u_j}({l_{j|{i,b(t)}}})f_{i}(t)dt}+
    (1-F_i(t_{w|i})){u_j}({l_{j|{i,nb}}}) )),
\end{equation}

where $i \in \{ {wr,wn,cn}\}$ and $j \in \{ {al,mc}\}$.

To this end, we can solve this optimization problem and obtain the decision with minimal economic risk.
\section{Numerical Experiments} \label{sec5}
In this section, we demonstrate the maintenance planning model by applying it to numerical experiments. Note that we choose numerical experiments instead of field experiments considering that autonomous trucks are not in use yet in the real world. The case scenarios are inspired by Swedish freight transport practices and set as intercity delivery where the truck mostly runs on the highway. Note that the proposed maintenance planning model applies to any road type, and the highway scenario is used in this paper for better heuristic interpretation. Then we conduct sensitivity analyses on model parameters and discuss practical issues of applying the model in the end.

\subsection{Basic implementation} \label{sec5.1}
In this subsection, we first describe the case scenario and the values of parameters in the model. Then we implement the model by simulating the case in Matlab. In the end, we investigate the performance of the proposed model on reducing economic risk.

\subsubsection{Parameter configuration} \label{sec5.1.1}
\begin{figure}[ht]
\centering
\includegraphics[width= .6\linewidth]{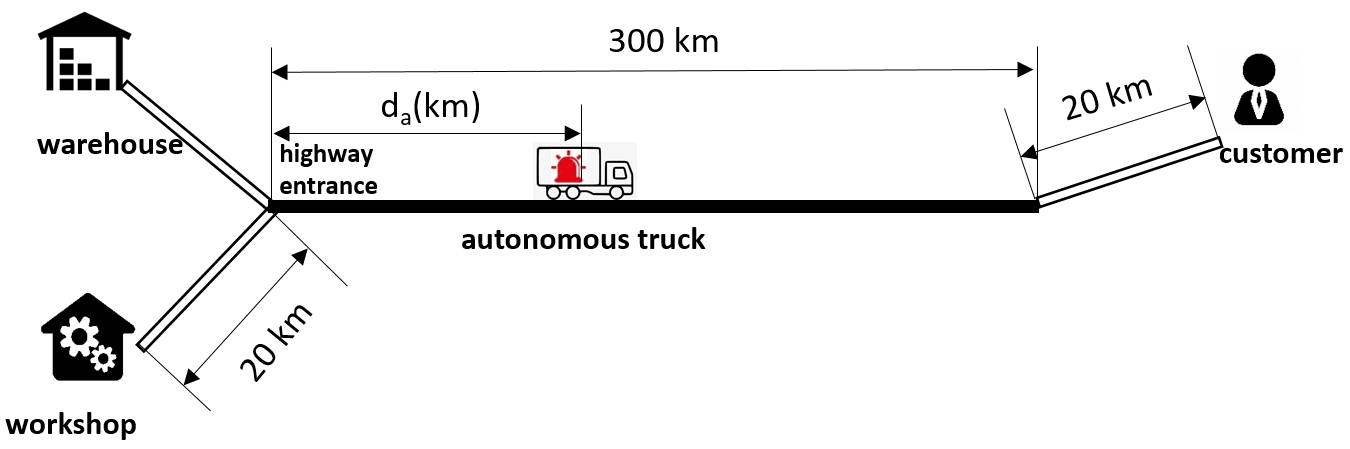}
\caption{Case scenario of the fright delivery mission with spatial information.}
\label{fig:5}
\end{figure}

As modelled in Section \ref{sec4.2}, the decision impact assessment involves the the locations and distances of system actors. In the basic implementation, the locations of the workshop and customer are fixed with spatial information shown in Fig. \ref{fig:5}. In this scenario, there is one available workshop in the city where the warehouse locates. The autonomous truck is loaded with goods and drives on a highway to the customer. On the highway, the truck encounters a fault alarm. The alarm location is represented by the distance in kilometer (km) from the alarm location to the highway entrance $d_a $. The economic risks of different decisions with different alarm locations are computed with $d_a \in \{0, 1, 2, ..., 300\}$.

In \textbf{Table 1}, we list the parameter value related to driving speed, maintenance time, and maintenance cost in the maintenance planning model. 

Furthermore, we assume that the truck company reaches an agreement with the customer on the economic utility of availability loss $u_{al}(l_{al})$, which can be presented by a piece-wise linearly function and presented in Eq. (\ref{eq:24}). \\

\renewcommand{\arraystretch}{1.2}
\noindent\begin{tabular} {p{2 cm} p{2 cm} p{2 cm}}
\multicolumn{3}{l}{\textbf{Table 1}}\\
\multicolumn{3}{l}{Parameter value related to speed, time and cost.}\\\hline
parameter    & value & unit    \\\hline
$v_n$        & 80    & km/hour \\ 
$v_{ul}$     & 80    & km/hour \\  
$v_{l}$      & 30    & km/hour \\ 
$v_{wr}$     &  40   & km/hour \\ 
$v_{wn}$     &  80   & km/hour \\ 
$v_{cn}$     &  80   & km/hour \\ 
$c_{fixed}$  & 75    & {EUR}   \\ 
$c_{var}$    & 2.5   & {EUR}/km\\
$c_{m|{nb}}$ & 500   & {EUR}   \\
$c_{m|{b}}$  & 1000  & {EUR}   \\ 
$t_{m|{nb}}$ & 2     & hour    \\
$t_{m|{b}}$  & 4     & hour    \\
$t_{sc}$     & 0.5   & hour    \\
\hline
\label{tab1} 
\end{tabular}

\begin{equation} \label{eq:24}
    u_{al}(l_{al}) = \left\{ \begin{array}{l}
0,\quad{\rm{  if  }}\quad l_{al} \in [0, t_{min}];\\
  \\
100(l_{al}-t_{min}),\quad{\rm{   if  }}\quad l_{al} \in (t_{min},t_{max}];\\
  \\
{pe}_{max},\ \ {\rm{ if  }}\quad l_{al} \in (t_{max}, + \infty ).
\end{array} \right.
\end{equation}

In Eq. (\ref{eq:24}), availability loss $l_{al}$ is scaled as the loss of time. A maximal time of delivery delay $t_{min}$ is allowed without penalty; the time of delivery delay between $t_{min}$ and $t_{max}$ should be penalized with 100 EUR/hour to the customer; when the time of delivery delay is more than $t_{max}$, the order will be cancelled and a total penalty of ${pe}_{max}$ is charged. In the basic implementation, these parameters are selected as $t_{min}=2$ hours, $t_{max}=10$ hours, ${pe}_{max}= 2000$ EUR. Note that utility function $u_{al}(l_{al})$ could be different, while the one in Eq. (\ref{eq:24}) is inspired by contracts in the truck business. 

Since maintenance cost is a direct economic loss, the economic utility of maintenance cost is equal to this cost:
\begin{equation} \label{eq:25}
    u_{mc}(l_{mc})=l_{mc}.
\end{equation}

For the distribution of RUL, we follow \cite{van2009survey} and use a Gamma function to describe the degradation process of the faulty component. The RUL of the faulty component given decision $i$, denoted as $T_i$, is a random variable following Gamma distribution. The probability density function of RUL, $f_i(t)$ is shown in Eq. (\ref{eq:26}) and the cumulative density function $F_i(t)$ is shown in Eq. (\ref{eq:27}): 
\begin{equation} \label{eq:26}
    f_i(t) = \frac{1}{{{\beta_i ^{\alpha_i} }\Gamma (\alpha_i )}}{t^{\alpha_i  - 1}}{e^{\frac{{ - t}}{\beta_i }}},
\end{equation}

\begin{equation} \label{eq:27}
    F_i(t) = \frac{1}{{{\beta_i ^{\alpha_i} }\Gamma (\alpha_i )}}\int_0^t {{t^{\alpha_i  - 1}}{e^{\frac{{ - t}}{\beta_i }}}} dt,
\end{equation}
where $\alpha_i$ is the shape parameter, $\beta_i$ is the scale parameter, and $ \Gamma (\alpha_i )$ is a gamma function. For all positive integers,  $ \Gamma (\alpha )= (\alpha -1)!$.

\begin{figure}[ht]
\centering
\includegraphics[width=.6\linewidth]{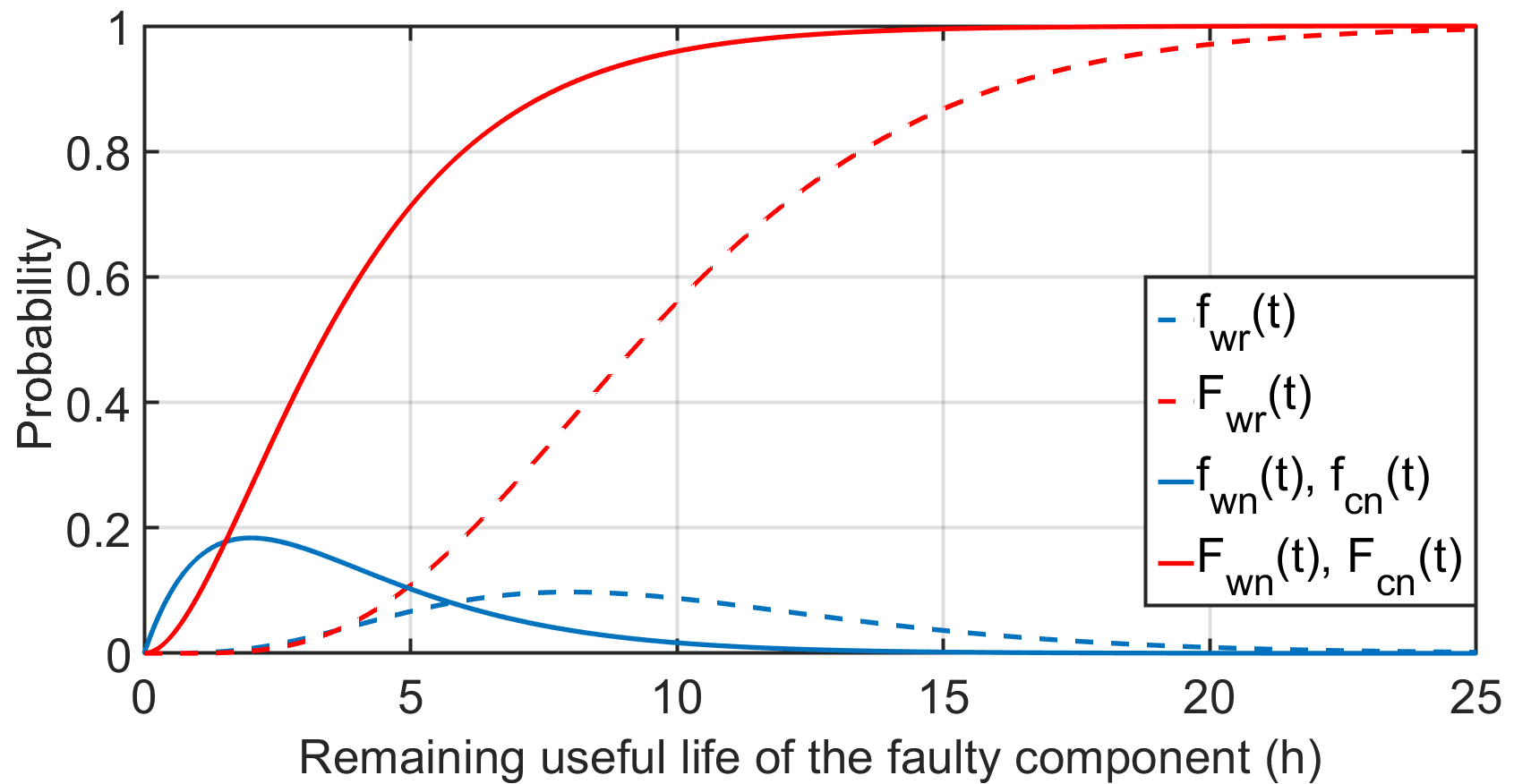}
\caption{Probability density function and cumulative probability function of the remaining useful life of the faulty component with different decisions.}
\label{fig6}
\end{figure}

In this basic implementation, the selected values of
parameters $\alpha_i$ and $\beta_i$ are listed in \textbf{Table 2}. The probability densities $f_i(t)$ and and cumulative probabilities $F_i(t)$ are shown in Fig. \ref{fig6}. In this figure, the RUL by taking decision $wr$ (dashed blue line) is much longer compared to that by taking decision $wn$ or $cn$ (solid blue line). This corresponds to the assumption in Section \ref{sec4.1} that a reduced driving speed can prolong the RUL of the faulty component. \\

\renewcommand{\arraystretch}{1.2}
\noindent\begin{tabular}{p{2 cm} p{2 cm} p{1 cm} p{1 cm}}
\multicolumn{4}{l}{\textbf{Table 2}}\\
\multicolumn{4}{l}{Parameters of probability density functions.}\\\hline
 Decision i & $f_i(t)$ & $\alpha_i$ & $\beta_i$ \\ \hline
 $wr$ & $f_{wr}(t)$ & $5 $ & $2 $ \\ 
 $wn$ & $f_{wn}(t)$ & $2 $ & $2 $ \\ 
 $cn$ & $f_{cn}(t)$ & $2 $ & $2 $ \\ 
 \hline
\label{tab2}
\end{tabular}

\subsubsection{Implementation results} \label{sec5.1.2}
\begin{figure}[ht]
\centering
\includegraphics[width=.6\linewidth]{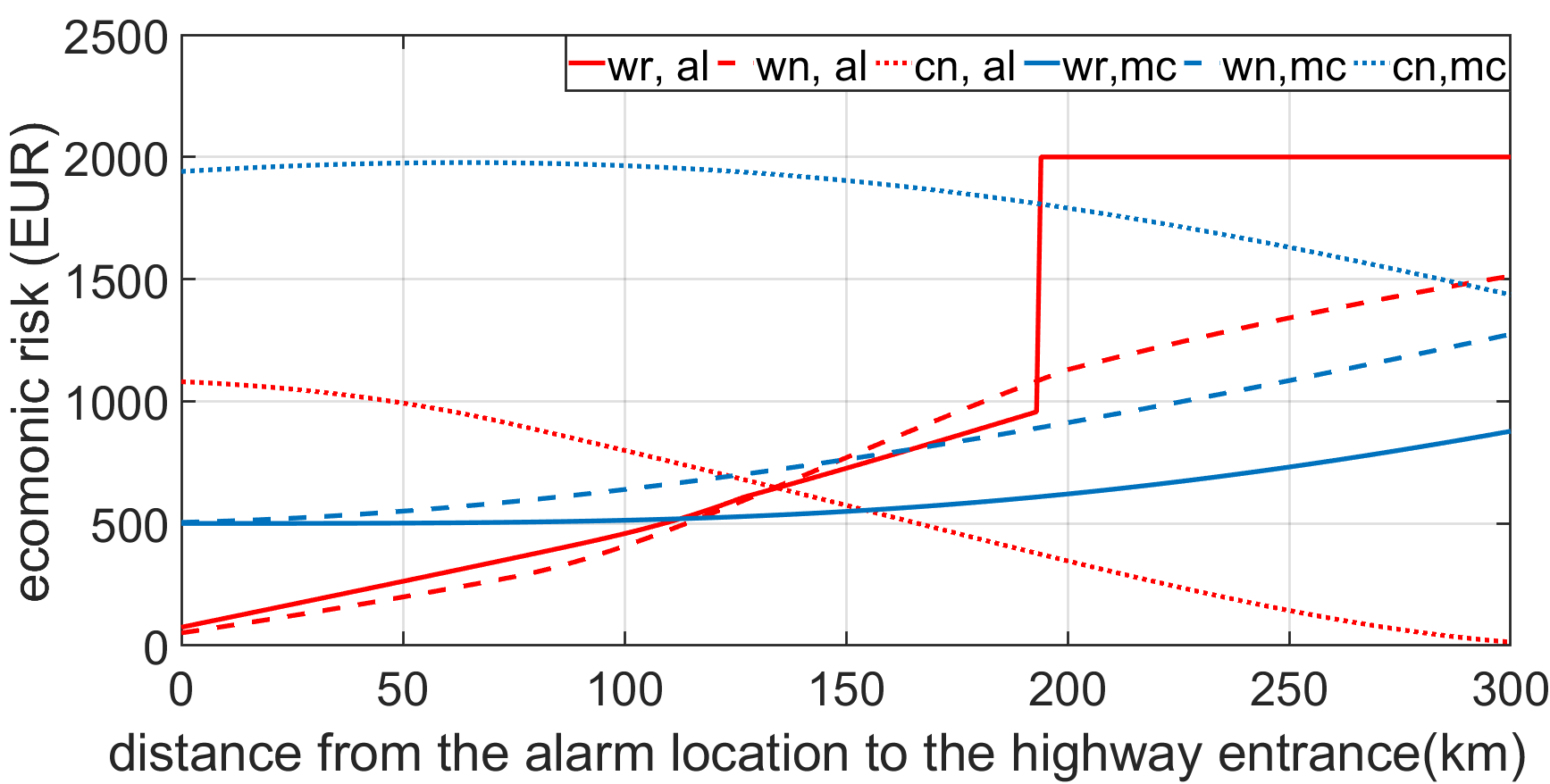}
\caption{Economic risk of availability loss and maintenance cost of different decisions at different alarm locations.}
\label{fig7}
\end{figure}

\begin{figure}[ht]
\centering
\includegraphics[width=.6\linewidth]{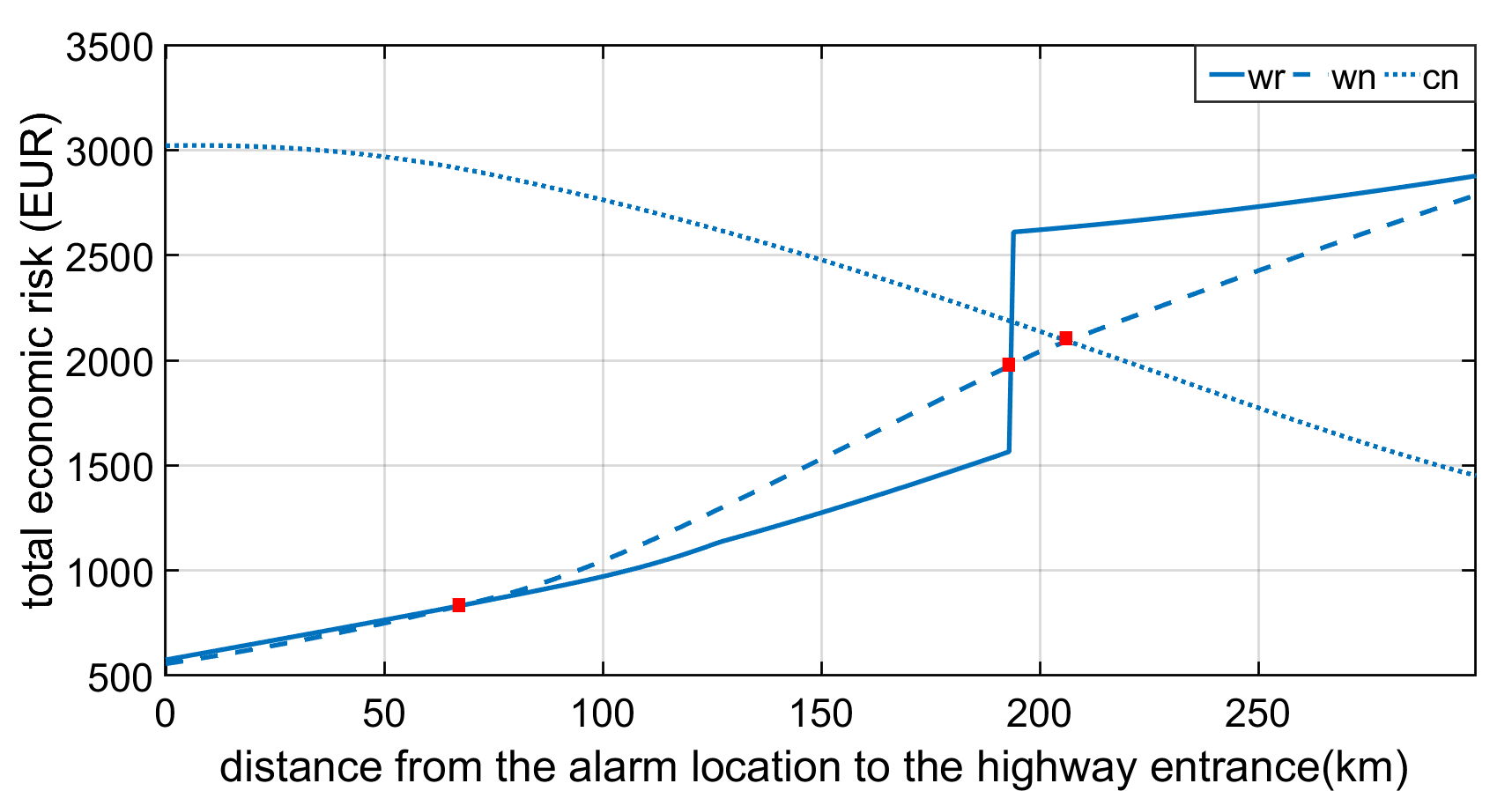}
\caption{Total economic risk of different maintenance decisions at different alarm locations.}
\label{fig8}
\end{figure}

The proposed maintenance planning model is configured and applied to the case scenario by simulation in Matlab. The economic risk of availability loss and maintenance cost given different decisions is shown in Fig. \ref{fig7}. The total economic risk given different decisions is shown in Fig. \ref{fig8}. The economic risk at different alarm locations is evaluated every 1km with a total highway distance of 300 km. 

In Fig. \ref{fig7}, the changes of economic risk have different trends. The economic risk of availability cost given decision $wr$ and $wn$ increases as $d_a$ increases. This is intuitive since it takes more time to reach the workshop when the alarm location is further from the workshop. However, the economic risk of availability loss of decision $cn$ decreases steadily. This is also intuitive since the truck is very likely to deliver the goods without a breakdown on the road when the alarm location is close to the customer. Note that there is a sharp increase for decision $wr$ at $d_a=190$ km. This corresponds to the situation that the availability loss exceeds the maximal allowed time $t_{max}$ and a large penalty $pe_{max}$ is charged. The correspondence of experiment results and intuition contributes to validating the effectiveness of the proposed method.

As shown in Fig. \ref{fig7}, the economic risk of availability loss and that of maintenance cost have similar magnitude. Therefore, neither of them should be ignored when making a maintenance decision. The decision output by considering the total economic risk (see Fig. \ref{fig8}) is different from that by only considering one decision impact (see Fig. \ref{fig7}). For example, at the alarm location $d_a=200$ km, the optimal decision is $cn$ if considering availability loss only, it is $wr$ if considering maintenance cost only, and it is $wn$ if considering both of them. Such a distinctive difference of decision outputs indicates the importance of considering both availability loss and maintenance cost.

To this end, given the alarm location ${d_a}$ and decision $i$, we can obtain the economic risk $\mathrm{E}[L_{i|d_a}]$. To evaluate the performance of the proposed maintenance planning method, we compare the proposed method to three baseline methods. Baseline method $i$, denoted as $bm_i$ is the method that always takes maintenance decision $i$. 

Given the distribution of alarm locations, the expected economic risk of baseline method $bm_i$, denoted as ${E\!E\!R}(bm_i)$, is computed as 
\begin{equation}
{E\!E\!R}(bm_i)= \mathrm{E}_{d_a}[\mathrm{E}[L_{i|d_a}]].
\end{equation}

The expected economic risk of the proposed method, denoted as ${E\!E\!R}(pm)$, is computed as 
\begin{equation}
{E\!E\!R}(pm)= \mathrm{E}_{d_a}[min(\mathrm{E}[L_{i|d_a}])].
\end{equation}

Assuming the alarm occurs randomly on the highway, i.e. ${d_a}\sim \mathrm{U}[0,300]$, the expected economic risk of different methods can be computed as
\begin{equation}
{E\!E\!R}(bm_i)= \sum_{d_a=0}^{300}[\mathrm{E}[L_{i|d_a}]]
\end{equation}
and
\begin{equation}
{E\!E\!R}(pm)= \sum_{d_a=0}^{300}[min(\mathrm{E}[L_{i|d_a}])].
\end{equation}

The expected economic risk values of different methods are listed in \textbf{Table 3}. \\ 

\renewcommand{\arraystretch}{1.2}
\noindent\begin{tabular}{ p{4 cm} p{2 cm}  } 
\multicolumn{2}{l}{\textbf{Table 3}}\\ 
\multicolumn{2}{l}{Expected economic risk of different methods (EUR).}\\ \hline
${E\!E\!R}(bm_{wr})$ & $1617$  \\ 
${E\!E\!R}(bm_{wn})$ & $1574$  \\ 
${E\!E\!R}(bm_{cn})$ & $2396$  \\ 
${E\!E\!R}(pm)$ & $1279$\\ \hline
\end{tabular} \\

To evaluate the performance of the proposed method, we construct an indicator $R_i$, the reduction ratio of the expected economic risk of the proposed method over that of baseline method $bm_i$, expressed as:

\begin{equation}\label{eq:28}
R_i= \frac{({E\!E\!R}(bm_i)-{E\!E\!R}(pm))}{{E\!E\!R}(bm_i)}.
\end{equation}

The values of $R_i$ are listed in \textbf{Table 4}. The expected economic risk of the proposed method reduces significantly ($47\%$) compared to decision $cn$. It also reduces notably compared to decision $wr$ ($21\%$) and decision $wn$ ($19\%$). Therefore, the proposed method can reduce the expected economic risk effectively.\\

\renewcommand{\arraystretch}{1.2}
\noindent \begin{tabular}{ p{2 cm} p{2 cm} p{2 cm} } 
\multicolumn{3}{l}{\textbf{Table 4}}\\ 
\multicolumn{3}{l}{Reduction ratio of expected economic risk $R_i$.}\\ \hline
$i={wr}$ & $i={wn}$ & $i={cn}$ \\
$21\%$   &  $19\%$  & $47\%$  \\ \hline
\label{tab4}
\end{tabular}
\subsection{Sensitivity analysis} \label{sec5.2}
In this subsection, we conduct a sensitivity analysis of model parameters, including the RUL of the faulty component, the economic utility of availability loss, and the number of workshops.

\subsubsection{Remaining useful life} \label{sec5.2.1}
The development of fault prognosis technology leads to a more accurate estimation of RUL, which can be presented by a smaller variance of the estimated RUL. Therefore, we conduct a sensitivity analysis of the variance of RUL. The variance and the expectation of the estimated RUL with a distribution $f_i(t)$ are
\begin{equation}\label{eq:29}
    \mathrm{Var}_i= \alpha_i{\beta_i}^2
\end{equation}
and
\begin{equation}\label{eq:30}
    \mathrm{E}_i= \alpha_i\beta_i.
\end{equation}

Note that in practice $\alpha_i$ and $\beta_i$ are provided by the prognosis module directly. In this experiment, we define them based on the two settings below. 

1) Keep variance as the only variable. As shown in Eq. (\ref{eq:29}) and Eq. (\ref{eq:30}), both the variance and the expectation relate to $\alpha_i$ and $\beta_i$. By only varying $\alpha_i$ or $\beta_i$, both the variance and the expectation change. Therefore, we set the expectation as a constant $\mathrm{E}_{wn}=4$ and sample 45 $\alpha_{wn}$ values uniformly between 1 and 10, i.e. $\alpha_{wn} \in \mathrm{U}(1,10]$. The corresponding $\beta_{wn}$ value is obtained by solving $\alpha_{wn}\beta_{wn}=\mathrm{E}_{wn}=4$. The parameters of $f_{cn}(t)$ are the same as that of $f_{wn}(t)$, i.e. $ {\alpha _{cn}}=\alpha _{wn} $, $\beta _{cn}=\beta _{wn} $.

2) The RUL given decision $wr$ is much longer than that given decision $wn$. This setting corresponds to the assumption in Section \ref{sec4.1} that reducing the driving speed can prolong the RUL of the faulty component. To model it and determine the parameters of $f_{wr}(t)$, we further assume that i) the expected distance the truck drives at a reduced speed before breakdown is twice of that at a normal speed, and {ii}) the variance of RUL given decision $wr$ is the same as that given decision $wn$. These two assumptions are expressed as the equation set: 
\begin{equation}\label{eq:31}
    \left\{ \begin{array}{l}
{\alpha _{wn}}\beta _{wn}^2 = {\alpha _{wr}}\beta _{wr}^2,\\
2{\alpha _{wn}}{\beta _{wn}}{v_{wn}} = {\alpha _{wr}}{\beta _{wr}}{v_{wr}}.
\end{array} \right.
\end{equation}

After obtaining $\alpha_{wn}$ and $\beta_{wn}$, we can obtain the corresponding $\alpha_{wr}$ and $\beta_{wr}$ by solving Eq.(\ref{eq:31}). 

Integrating $\mathrm{E}_{wn}=4$ assumed inthe first setting into Eq. (\ref{eq:29}), we obtain the relation between $\mathrm{Var}_{wn}$ $\alpha_{wn}$ as 
 \begin{equation}\label{eq:32}
     \mathrm{Var}_{wn} = {\alpha _{wn}}{\beta _{wn}}^2 = \frac{{16}}{{\alpha _{wn}}}.
 \end{equation}

As shown in Eq.(\ref{eq:32}), when $\alpha_{wn}$ increases, the variance $\mathrm{Var}_{wn}$ will decrease. Therefore, a small $\alpha_{wn}$ corresponds to a large $\mathrm{Var}_{wn}$, indicating a less accurate estimation of RUL. 

\begin{figure} [ht]
\centering
\includegraphics[width=.6 \linewidth]{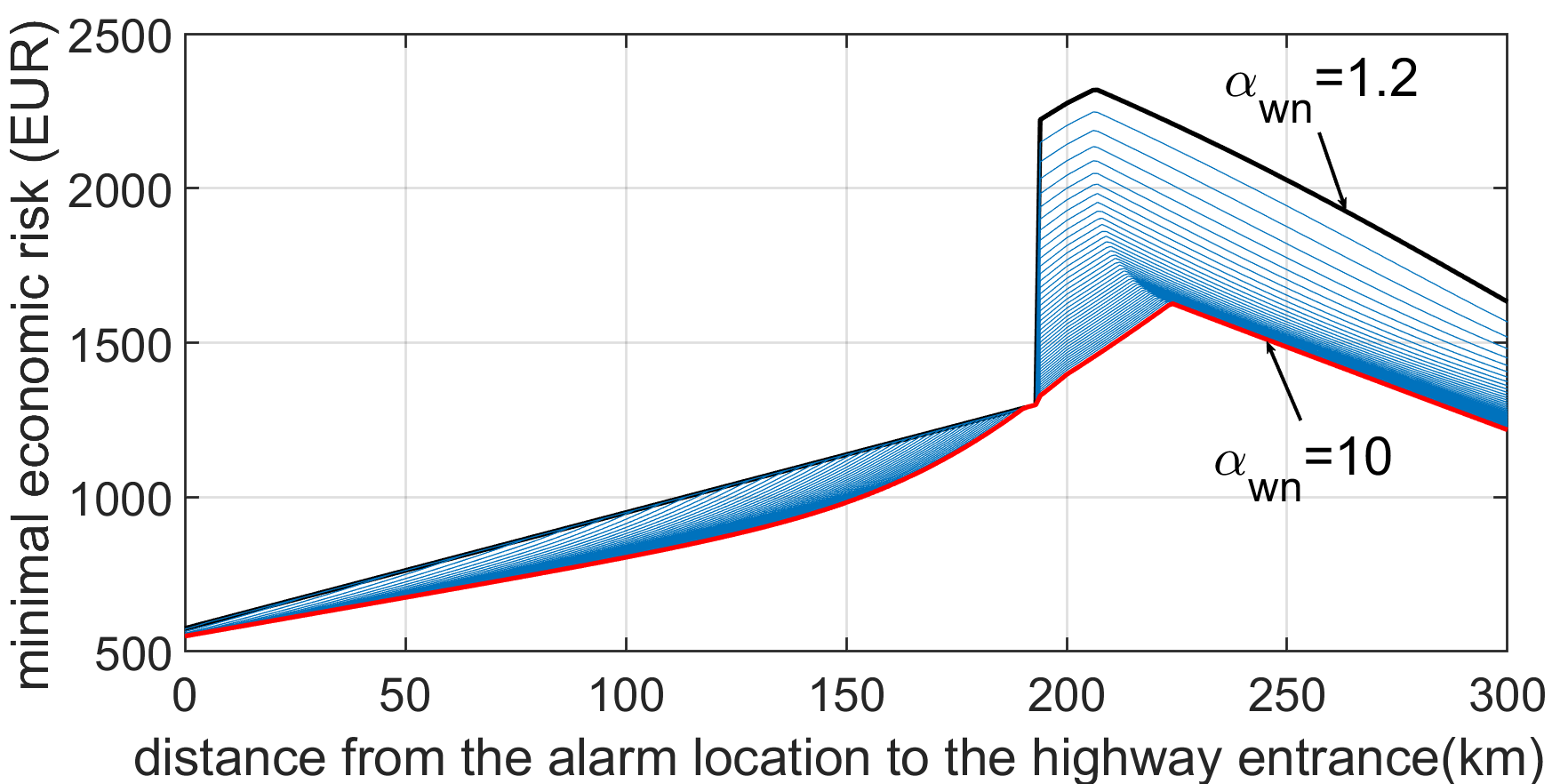}
\caption{The relation between the minimal economic risk and $\alpha_{wn}$ with $\alpha_{wn} \in \mathrm{U}(1,10]$.}
\label{fig9}
\end{figure}

With the same configurations in Section \ref{sec5.1.1}, the sensitivity result is shown in Fig. \ref{fig9}, illustrating the MER with different $\alpha_{wn}$. This MER is obtained with the proposed method. As shown in Fig. \ref{fig9}, as a whole, the smaller $\alpha_{wn}$ is, the higher the MER is. In other words, the less accurate the estimation of RUL is, the higher the MER is. 

\begin{figure} [ht]
\centering
\includegraphics[width=.6 \linewidth]{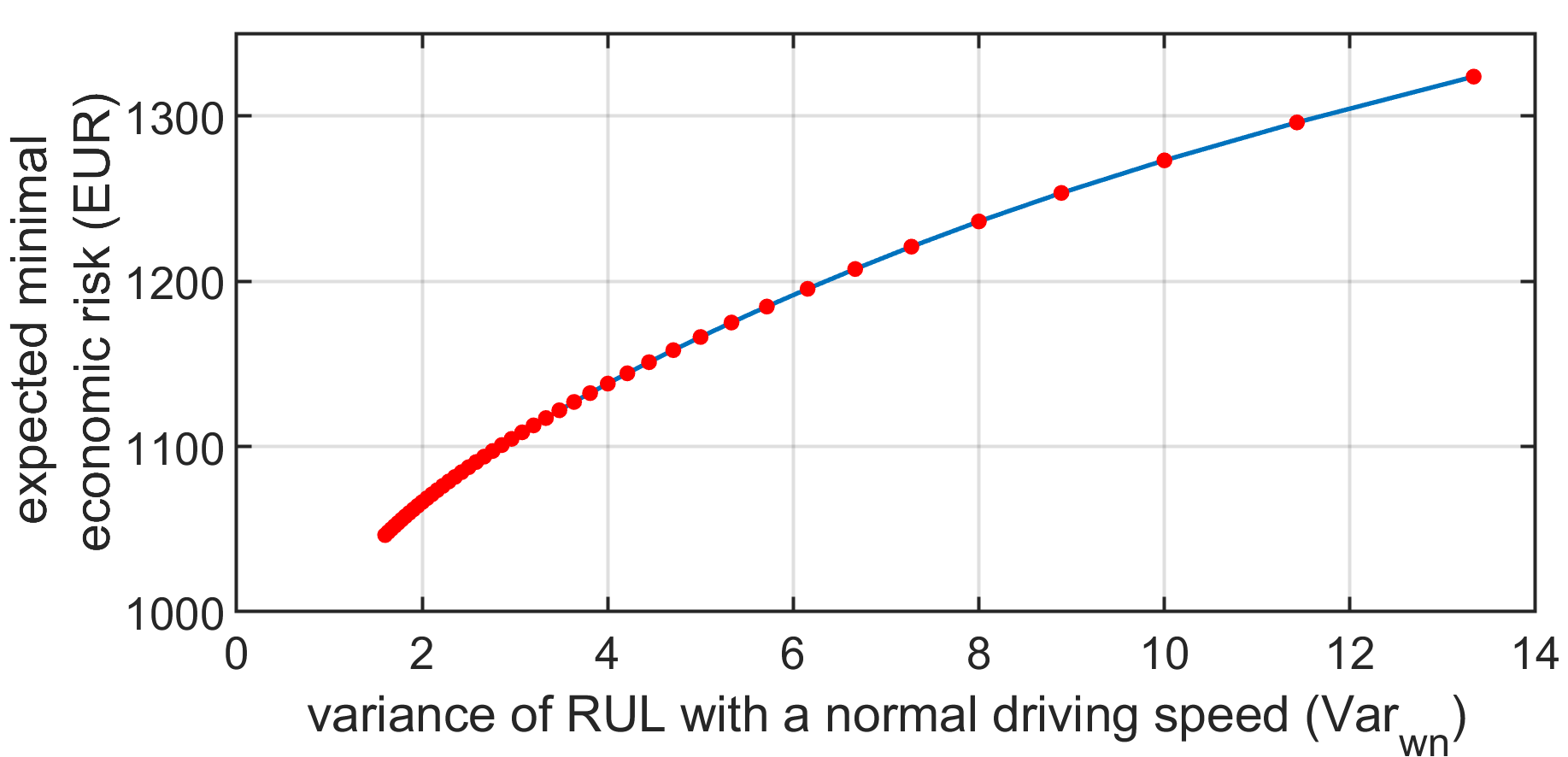}
\caption{The relation between the expected minimal economic risk and the variance of RUL with a normal driving speed.}
\label{fig10}
\end{figure}

The influence of the variance $\mathrm{Var}_{wn}$ on the expected MER is shown in Fig. \ref{fig10}. Assume that the fault occurs randomly on the highway, the expected  MER is obtained by taking the average of MER at different alarm locations. In Fig. \ref{fig10}, the expected MER increases steadily as $\mathrm{Var}_{wn}$ increases. This sensitivity analysis indicates that a more accurate RUL estimation can reduce the expected MER. 

\subsubsection{Economic utility of availability loss} \label{sec5.2.2}
As discussed in Section \ref{sec5.1.1}, the economic utility of availability loss depends on the agreement between the truck company and the customer, which is highly uncertain in practice. Therefore, we follow the agreement in Eq.(\ref{eq:24}) and conduct a sensitivity analysis on the maximal allowed time of delivery delay before order cancellation $t_{max}$ and the penalty of order cancellation ${pe}_{max}$. Specifically, two groups of samples are selected. The first group is configured as $t_{min}=2$ (hours), $t_{max}=10$ (hours), with 9 ${pe}_{max}$ values uniformly sampled in $[800,4000]$(EUR). The second group is configured as $t_{min}=2$ (hours), $t_{max}=6$ (hours), with 9 ${pe}_{max}$ values uniformly sampled in $[800,4000]$(EUR). 

\begin{figure} [ht]
\centering
\includegraphics[width=.6 \linewidth]{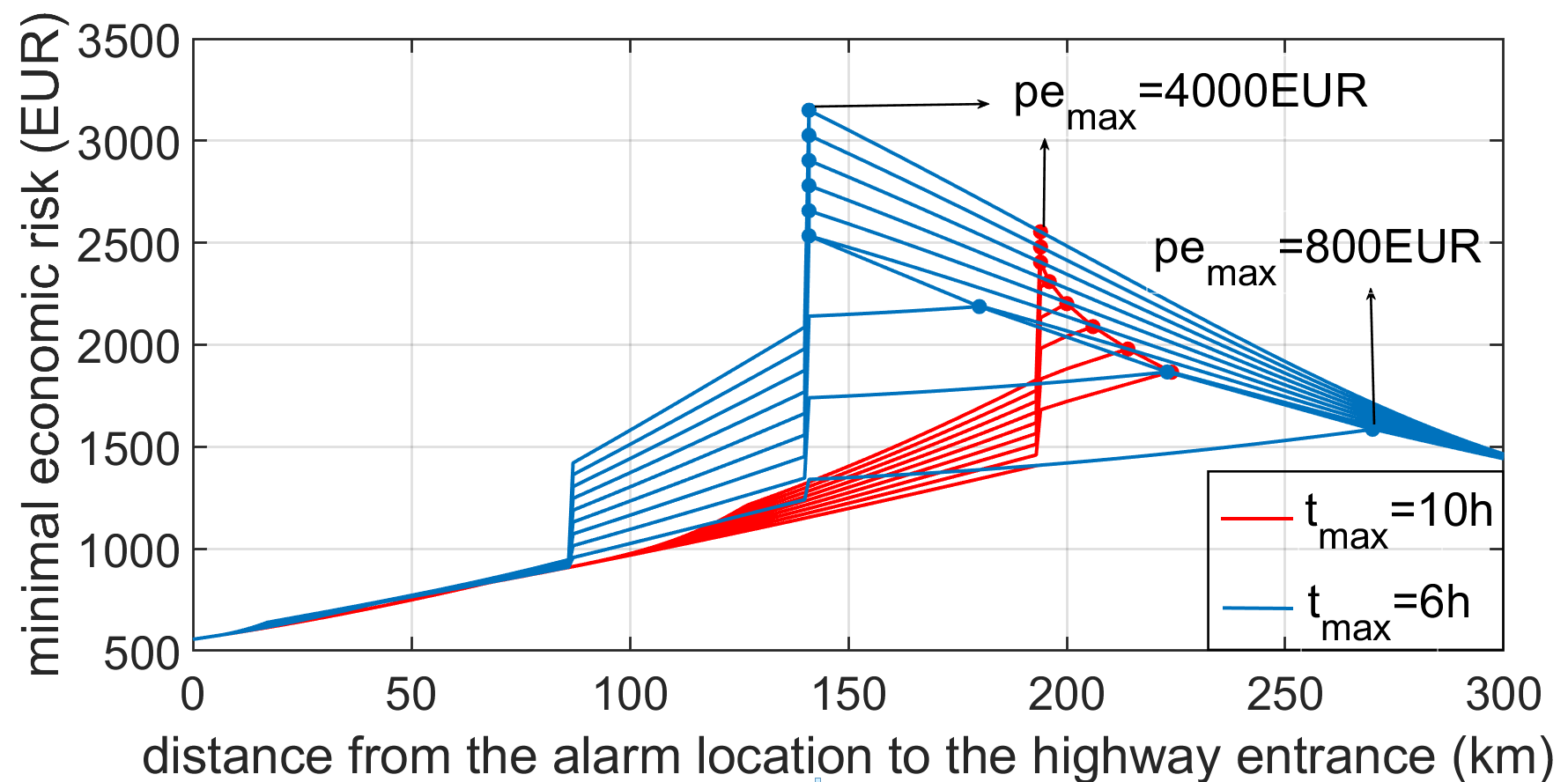}
\caption{Minimal economic risk with different utility functions of availability loss. Two variables in the utility function are varied, including the maximal allowed time of delivery delay before order cancellation $t_{max}$ and the penalty of order cancellation ${pe}_{max}$.}
\label{fig11}
\end{figure}

The simulation result is shown in Fig. \ref{fig11}. As shown, both $t_{max}$ and ${pe}_{max}$ influence the MER. Compared to a small $t_{max}$ (blue lines), with a large $t_{max}$ (red lines), the MER is relatively lower and ${pe}_{max}$ has smaller influence on the MER. Furthermore, assuming a random occurrence of fault on the highway, with a large ${pe}_{max}=4000$ EUR, the expected MER of $t_{max}=10$ hours reduces $20\%$ compared to that of $t_{max}=6$ hours. Therefore, a large $t_{max}$ makes MER more robust to ${pe}_{max}$ and reduces the expected MER effectively. 

\subsubsection{Two-workshop scenario} \label{sec5.2.3} 
\begin{figure}[ht]
\centering
\includegraphics[width=.6\linewidth]{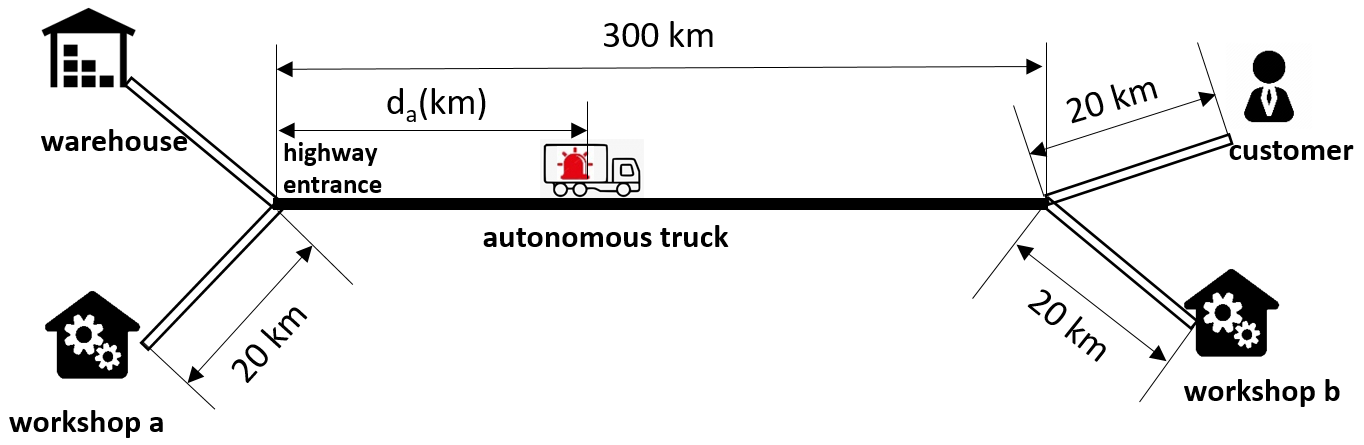}
\caption{Case scenario and spatial information with two available workshops.}
\label{fig12}
\end{figure}

In this subsection, the influence of the number of available workshops on the MER is explored, considering that more maintenance resources are available and connected with the development of communication technologies. Specifically, we assume that there are two available workshops, denoted as workshop a and b. The case scenario with spatial information is shown in Fig. \ref{fig12}. The distances are the same as in the basic implementation. 

\begin{figure}[ht]
\centering
\includegraphics[width=.6\linewidth]{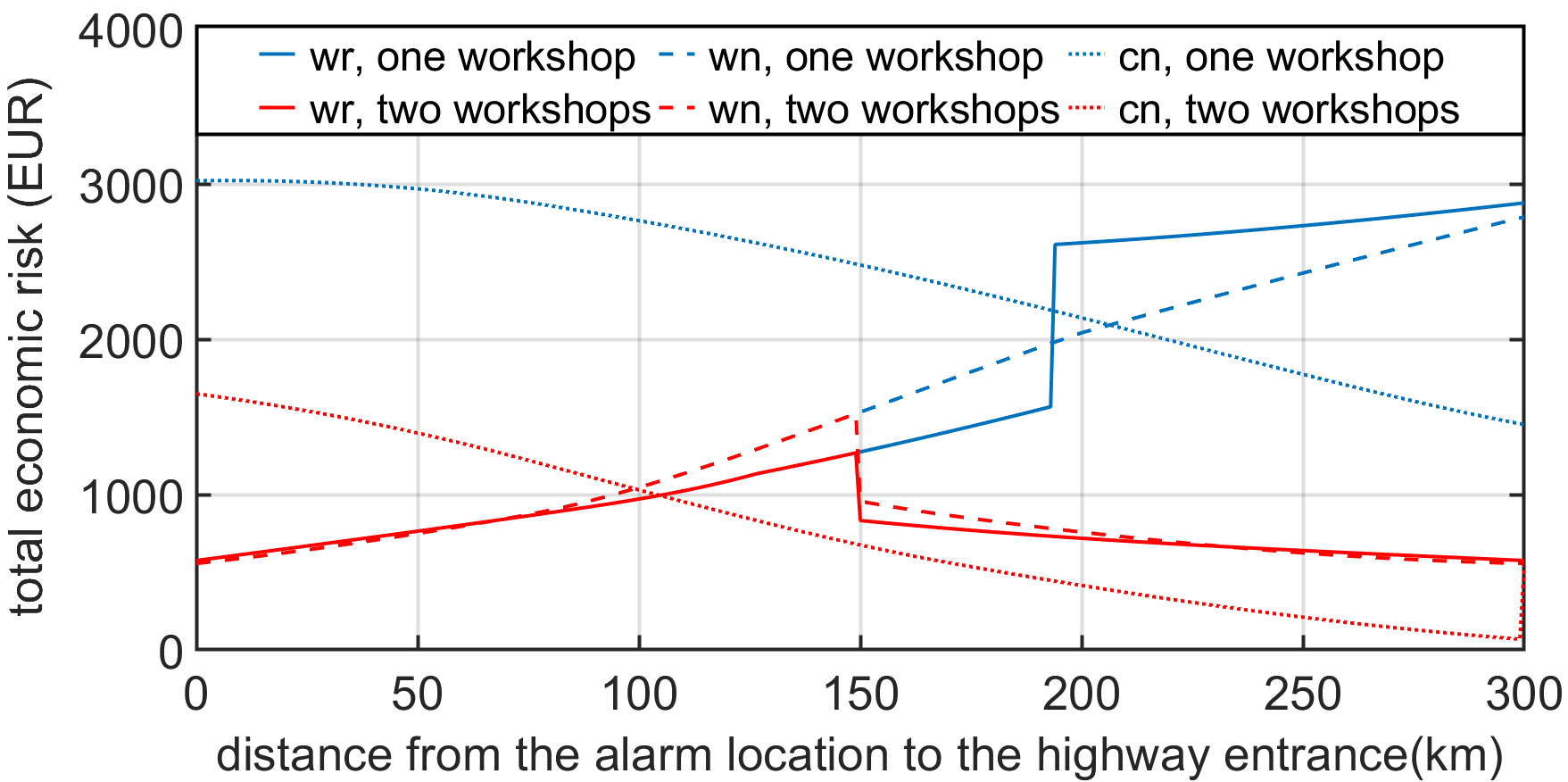}
\caption{Total economic risk of different maintenance decisions with one or two workshops.}
\label{fig13}
\end{figure}

The influence of adding one more workshop on the total economic loss of different decisions is shown in Fig. \ref{fig13}. With two available workshops, the total economic risk of decision $wr$ and $wn$ both drop significantly when $d_a>150$ km. The reason for the drop is that the nearest available workshop changes from workshop a to workshop b when $d_a>150$ km, so that the distance from the alarm location to the workshop decreases sharply. The total economic risk of decision $cn$ reduces to approximately half as the case with one workshop. Assuming a random occurrence of the fault alarm on the highway, the expected MER has a significant decrease of $57\%$, indicating that more workshop resources contribute to decreasing the expected MER.
\subsection{Discussion on practical issues} \label{sec5.3}
In the previous subsection, the numerical experiment results indicate the effectiveness and performance of the proposed method on a hypothetical example. However, issues may be encountered when applying this method in a real-world setting,  which are discussed below. 

\textbf{Maintenance decision alternatives.} As discussed in Section \ref{sec4.1}, we consider a collection of three decision alternatives in this paper, which represent three typical types, i.e. risk-aversion, risk-neutral, and risk-seeking ones. Each of these types contains more alternatives than the presented ones. For example, the risk-seeking type is represented by decision $cn$ considering immediate and unplanned maintenance after the delivery task. However, scheduled maintenance after delivery tasks also belongs to this type, which may cost much less money and time than an unscheduled one. Furthermore, depending on the fault types and available maintenance resources, maintenance intervention can be executed not only by workshops. For example, onsite maintenance by an assistant vehicle can be applied with some faults like a flat tire. Therefore, a more detailed classification of decision alternatives can help to evaluate the economic risk more accurately, which, however, is not the focus of this paper. 

\textbf{RUL of the faulty component.} In this paper, decision $wr$ considers driving at a reduced speed to the workshop, which is selected to represent a decision type, i.e. risk-aversi-on ones. With this decision, we assume that a reduced driving speed can prolong the RUL of the faulty component. In the real case, the RUL is estimated and provided by the prognosis module, and not necessarily correlated to the driving speed positively. Such a correlation needs more detailed modeling, which is not the scope of this paper. However, this assumption in the paper is reasonable for some types of faults, especially component degradation faults.

\textbf{Risk factors.} In this paper, we consider the risk of truck breakdown, which is not the only existing risk in practice. For example, in the context of autonomous driving, communication time is usually considered to be instantaneous. However, a poor network can cause information delay and cause availability loss as well. Furthermore, with the risk of truck breakdown on the road, the hazardous events can be more than the ones mentioned in the paper. For example, the truck breakdown can cause infrastructure damage, further reducing the truck profit. These risk factors are not the main focus and are not considered in this paper. Despite that it is difficult to quantify these risks, they do exist and we need to be aware of them in practice.

\section{Conclusion} \label{sec6}
In this paper, we propose a short-term maintenance planning model for autonomous trucks using a risk-based decision-making method. The model is demonstrated and the performance is discussed by applying it to numerical experiments. A sensitivity analysis is conducted of the RUL, economic utility of availability loss, and the number of available workshops. The experiment validates the necessity of considering both availability loss and maintenance cost, as well as the effectiveness of the proposed method in reducing economic risk. The sensitivity analysis shows that the minimal economic risk decreases significantly by increasing the estimation accuracy of RUL, the maximally allowed availability loss, or the number of available workshops. Furthermore, by increasing the maximally allowed availability loss, the economic risk is more robust to an increase in the penalty of order cancellation. These sensitivity analysis results contribute to identifying future research and development attentions of autonomous trucks from an economic perspective.

While this paper is focused on building a short-term maintenance planning model for autonomous trucks, more efforts are needed to increase the generalization and applicability of the model in practice. The discussion on practical issues in Section \ref{sec5.3} indicates potential improvements by considering more decision alternatives, a more accurate RUL model, and more risk factors. Furthermore, the proposed maintenance planning model can be further generalized to deal with more complex scenarios, such as autonomous truck fleets and autonomous trucks with multiple consecutive delivery tasks.\\

\noindent
\textbf{Acknowledgement}\\
This research was supported by the KTH-CSC Programme, funded by the China Scholarship Council (CSC) (grant no. 201700260189).

\bibliographystyle{cas-model2-names}

\bibliography{sample}

\end{document}